\documentclass[11pt]{article}

\usepackage[preprint]{acl}

\usepackage{paralist}
\usepackage{times}
\usepackage{latexsym}
\usepackage[T1]{fontenc}
\usepackage[utf8]{inputenc}
\usepackage{microtype}
\usepackage{inconsolata}
\usepackage{graphicx}
\usepackage{amsmath}
\usepackage{cuted}
\usepackage{capt-of}
\usepackage{hyperref}
\usepackage{amsfonts}
\usepackage{booktabs}
\usepackage{tabularx}
\usepackage{array}
\usepackage[table]{xcolor}
\usepackage{listings}
\usepackage{pifont}
\newcommand{\cmark}{\ding{51}}
\newcommand{\xmark}{\ding{55}}
\usepackage{amssymb} 
\usepackage{afterpage}
\usepackage{booktabs}
\usepackage{mathtools}
\usepackage{cleveref}
\usepackage{xspace}

\newcommand{\ex}[1]{\textit{#1}\xspace}
\newcommand{\ordval}[1]{\ensuremath{\mathrm{#1}}\xspace}
\newcommand{\high}{\ordval{high}}
\newcommand{\medium}{\ordval{medium}}
\newcommand{\low}{\ordval{low}}

\title{Beyond the Resum\'e: A Rubric-Aware Automatic Interview System for Information Elicitation}

\author{
Harry Stuart\thanks{Corresponding~author.} \qquad
Masahiro Kaneko \qquad
 Timothy Baldwin \\
Mohamed bin Zayed University of Artificial Intelligence (MBZUAI) \\
Abu Dhabi, United Arab Emirates \\
\texttt{\{harry.stuart,masahiro.kaneko,timothy.baldwin\}@mbzuai.ac.ae}
}

\begin{document}
\maketitle

\begin{abstract}

Effective hiring is integral to the success of an organisation, but it is very challenging to find the most suitable candidates because expert evaluation (e.g.\ interviews conducted by a technical manager) are expensive to deploy at scale. Therefore, automated resume scoring and other applicant-screening methods are increasingly used to coarsely filter candidates, making decisions on limited information. We propose that large language models (LLMs) can play the role of subject matter experts to cost-effectively elicit information from each candidate that is nuanced and role-specific, thereby improving the quality of early-stage hiring decisions. We present a system that leverages an LLM interviewer to update belief over an applicant's rubric-oriented latent traits in a calibrated way. We evaluate our system on simulated interviews and show that belief converges towards the simulated applicants' artificially-constructed latent ability levels. We release code, a modest dataset of public-domain/anonymised resumes, belief calibration tests, and simulated interviews, at \href{https://github.com/mbzuai-nlp/beyond-the-resume}{https://github.com/mbzuai-nlp/beyond-the-resume}. Our demo is available at \href{https://btr.hstu.net}{https://btr.hstu.net}.

\end{abstract}

\afterpage{
  \begin{figure*}[t]
    \centering
    \includegraphics[width=\textwidth]{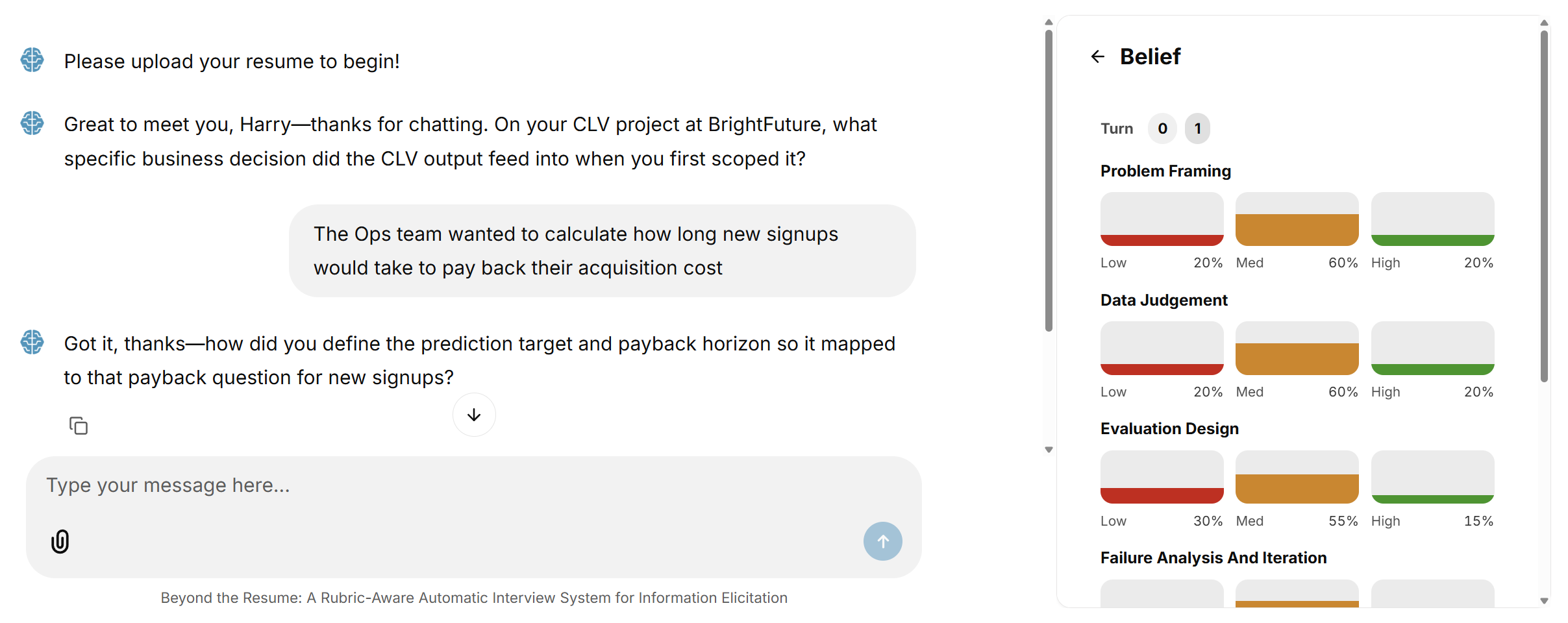}
    \caption{Screenshot of a user interacting with the system.}
    \label{fig:hero}
  \end{figure*}
}

\section{Introduction}

Hiring is a resource-intensive process and thus large pain-point for many organisations \cite{BhargavaAssadi2024Hiring, MuehlemannStruplerLeiser2018HiringCosts, BLATTER201220}. Ideally, subject matter experts (SMEs) such as senior employees from the relevant department would be engaged to conduct an initial interview with every candidate to elicit the specific, nuanced, role-specific information typically absent from a resume \cite{Risavy2022Resumes, Vincent_Guidice_Mero_2024}. In practice, this is cost-prohibitive.

Therefore, organisations increasingly employ screening procedures to reserve such hiring evaluations for only suitable candidates. Screening is typically carried out through automated resume scoring \cite{Li2020CompetenceLevel, 8117003} or interviews conducted by personnel who lack the expertise to perform high-cost, highly-informative evaluation \cite{10.1145/3544548.3581548}. Consequently, early hiring decisions are made in the absence of the information that matters most. This leads to the screening out of genuinely suitable candidates and the progression of unsuitable candidates \cite{Fisher2021Finding}. Downstream interviews are also less productive than if this supplemental, expertly-elicited information were made available beforehand.

We contend that large language models \cite[LLMs;][]{zhao2025surveylargelanguagemodels} can act as a low-cost proxy of SMEs during the early stages of hiring. Specifically, we introduce an open-source automatic multi-turn interview system that incorporates belief over a rubric representing the role-specific information a human SME would wish to elicit. \Cref{fig:hero} shows a snippet of the system mid-interview, with the rubric-belief in the right-panel.

We explore how to calibrate belief updates and then present belief convergence as a useful operationalisation of saturating a rubric with meaningful information about an applicant. We evaluate our system on simulated interviews and show that: (1) turn-to-turn belief change drops by roughly $3\times$ over the course of an interview on average; and (2) in simulation, where each applicant is generated from a fixed latent rubric profile (an ``archetype''), the final belief recovers the correct archetype $76.1\%$ of the time. 

An interview using our system produces two useful artifacts that enable more informed decision-making: (1) an information-rich transcript; and (2) a fully auditable log of belief updates. There is emerging research around using LLMs for recruitment interviews \cite{hashimoto-etal-2025-career, jokoarticle, adeseye2025modular} but existing systems largely frame elicitation as field collection or conversational scaffolding, rather than maintaining an explicit, calibrated probabilistic belief state over rubric-aligned latent attributes.

\section{System Overview}

\begin{figure}[t]
    \centering
    \includegraphics[width=\columnwidth]{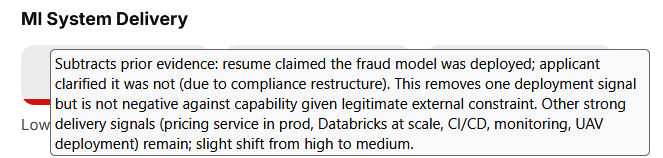}
    \caption{Judge's justification for updating belief over $\mathrm{ML\ Systems\ Delivery}$ dimension, in \textit{debasing} (diminishing the supportive power of) previous evidence. Not derived from the same interview presented in \Cref{fig:hero}.}
    \label{fig:justification}
\end{figure}

The demo of our system is deployed as a web-application (\Cref{fig:hero}) and assumes the hiring context of machine learning engineering. Our code repository contains instructions for customisation, self-hosting, and examples of other rubrics. The system runs a Python application using the Chainlit library \cite{chainlit-maintainers} to orchestrate the interview through a familiar LLM-chat interface. The belief panel on the right hand side presents the system's belief over each rubric dimension at each turn, with a turn selector to inspect previous belief. Additionally, belief changes can be audited as seen in \Cref{fig:justification}. This belief panel is only presented to the user for the purpose of this demonstration. In a real deployment scenario, belief would be hidden from the applicant and the interview transcript and belief log would be provided to the hiring team, likely via a dashboard or integration with existing hiring software.

Under the hood, our system comprises two components: (1) a \textit{judge}, implemented as an LLM, that tracks belief over a rubric; and (2) an \textit{interviewer}, also implemented as an LLM, whose objective is to elicit meaningful information from the applicant to saturate the rubric. Information that is meaningful provides evidence of the applicant's latent \textbf{Knowledge, Skills, and Abilities (KSA)} pertinent to the rubric \cite{doi:10.1177/009102609302200405, https://doi.org/10.1111/j.1744-6570.2000.tb00220.x, Hlavac02012023}. These respectively capture: what the applicant knows, what they can do, and how they behave. To saturate the rubric means to elicit information from the applicant such that the judge becomes confident in its belief.

All LLM components of this study use GPT5 \cite{openai_gpt5_release_2025}, specifically \texttt{gpt-5-2025-08-07}, with \texttt{reasoning="low"} to reduce token usage.

\subsection{Defining a Rubric}
\label{sec:rubric-def}

We define a rubric as having $D$ dimensions (e.g.\ $\mathrm{autonomy}$ or $\mathrm{programming\ ability}$), each with $L$ ordinal levels. Let $\mathcal{L}=\{1,\dots,L\}$ denote the set of level indices, ordered from lowest to highest. We represent the rubric as:
\begin{align}
R &=
\begin{bmatrix}
r_{1,1} & r_{1,2} & \cdots & r_{1,L} \\
\vdots  & \vdots  & \ddots & \vdots \\
r_{D,1} & r_{D,2} & \cdots & r_{D,L}
\end{bmatrix}.
\end{align}

Each $r_{d,\ell}$ is defined by a natural-language \textit{anchor} that describes the KSA entailed by level $\ell$ of dimension $d$ (e.g.\ $r_{\mathrm{autonomy},1}$ might be \ex{relies on others when problems are encountered; does not clarify initial requirements ...}). We define the following anchor requirements: 

\begin{compactitem}
    \item \textbf{Non-evidentiary:} The anchor exists independently of evidence provided by the applicant.
    \item \textbf{Falsifiable:} The anchor describes inferable and determinable KSA, avoiding strongly subjective qualifiers.
    \item \textbf{Internally-consistent:} within an anchor, described KSA should be jointly plausible for the same level, rather than mixing orthogonal KSA that may not co-occur.
\end{compactitem}

\subsection{The Judge}

We define a judge as an LLM policy that initialises and updates belief of the applicant's KSA. It initialises belief using the applicant's provided resume and updates its belief after each interview turn. Belief here is defined as a posterior distribution over the latent level of each rubric dimension, conditioned on all evidence observed so far (resume + interview transcript). The judge is also instructed to justify its belief updates so they are human-auditable.

Let $z_d \in \mathcal{L}$ denote the applicant's latent level for rubric dimension $d \in \{1,\dots,D\}$, and let $x_{0:t}$ denote the evidence available up to interview turn $t$ (with $x_0$ being the resume, and $x_{1:t}$ the interview turns). The judge's belief state at turn $t$ is:
\begin{align}
B_t = \left\{ p(z_d \mid x_{0:t}, R) \right\}_{d=1}^{D}.
\end{align}

\subsection{The Interviewer}
\label{sec:interviewer-intro}

We define an interviewer as a LLM policy whose objective is to, as previously described, elicit meaningful information from the applicant to saturate the rubric. We operationalise convergence as the mean reduction in the total variation (TV) distance between successive per-dimension posteriors:
\begin{align}
\Delta_t
=
\frac{1}{D}\sum_{d=1}^D
\mathrm{TV}\!\left(
[B_t]_d,\,
[B_{t-1}]_d
\right),
\label{eq:successive-tv}
\end{align}
where $\mathrm{TV}(p,q)=\tfrac12\sum_{\ell \in \mathcal{L}} |p(\ell)-q(\ell)|$ for distributions over $\mathcal{L}$. The ideal limiting behaviour is that, as informative evidence accumulates, $B_t$ concentrates on the true latent level $z_d$ for each dimension $d$. Convergence can be thought of as the judge becoming more confident in the hypotheses implied by its posteriors as they are tested under more information.

\section{Calibrating the Judge}

\subsection{Dataset}
\label{sec:judge-test-dataset}

In the absence of oracle data describing what it means to update belief, we define a set of tests that allow us to make assertions about the properties and behaviours of a judge. We begin by constructing a general rubric for each of these three orthogonal resume domains: graphic design, sales, and machine learning engineering. For experimental simplicity, all rubrics are instantiated with $L=3$ ordinal levels, with the identification $\mathcal{L}=\{1,2,3\}\equiv\{\low,\medium,\high\}$. For each resume domain, we download 10 public domain, anonymised resumes (for 30 total) from Reddit’s r/Resumes subreddit,\footnote{\url{https://www.reddit.com/r/Resumes/}} and extract the text content. For each resume, we manually construct a short interview, $\mathcal{I}$, of 2--5 turns to act as a stub in subsequent tests.

\subsection{Metamorphic Testing}

Metamorphic testing, borrowed from software engineering, specifies how an LLM's outputs
should change under controlled input perturbations without requiring oracle labels
\cite{Cho_2025, 10638599}.

Empirically, GPT5 tends to shift at least $0.05$ probability mass about a level given clear new evidence. We therefore treat a shift of at least $\varepsilon \coloneqq 0.04$ as a belief change. Define, for domain $d$ and level $\ell$,
\begin{equation*}
\begin{aligned}
p_d^{-}(\ell) &\coloneqq p(z_d=\ell \mid x_{0:t-1},R),\\
p_d^{+}(\ell) &\coloneqq p(z_d=\ell \mid x_{0:t},R),\\
\delta_d(\ell) &\coloneqq p_d^{+}(\ell)-p_d^{-}(\ell).
\end{aligned}
\end{equation*}
\begin{subequations}\label{eq:belief-updates}
\begin{align}
\mathcal{R}_{\mathrm{inv}} &:\ \max_{d,\ell} |\delta_d(\ell)| \le \varepsilon,
\label{eq:r-inv}\\
\mathcal{R}_{\uparrow}(d,\ell) &:\ \delta_d(\ell) > \varepsilon,
\label{eq:r-up}\\
\mathcal{R}_{\downarrow}(d,\ell) &:\ \delta_d(\ell) < -\varepsilon.
\label{eq:r-down}
\end{align}
\end{subequations}

We construct a test class for each of these three behaviours. A test within a class only passes if the corresponding behaviour is observed under the injection of a controlled interviewer--applicant turn, $\tau$. For each test class, we construct one $\tau$ per resume. We define two tests per resume by defining two input regimens: (1) $\text{resume-only}$; and (2) $\text{resume} \cup \mathcal{I}$. We use these two regimens to observe the judge's behaviour on the first turn of an interview and later in an interview. Therefore, we have a total of 60 tests per test class (30 resumes and two input regimens each), and nine test classes, making for 540 metamorphic tests in total. We present each test class in \Cref{tab:test-pass-rate-by-judge} beneath its expected relation. To \textit{debase} evidence means to diminish its supportive power. Examples and stratifications are provided in \Cref{sec:judge-strat}.

\subsection{Uniform Prior.}

In the absence of any information pertinent to a given rubric dimension, a judge should adopt a uniform belief across all levels. We test this by taking each resume and its corresponding rubric, then adding an unrelated dimension from one of the other rubrics and measuring whether the posterior over the \emph{levels of the injected dimension} is close to uniform.

Let \(d_{\mathrm{inj}}\) denote the injected unrelated dimension, and let
\(p_{d_{\mathrm{inj}}}\in\Delta^2\) be the posterior over its three levels.
This test checks whether that posterior is approximately uniform under an
\(L_\infty\) tolerance:
\begin{align}
\|p_{d_{\mathrm{inj}}}-(\tfrac13,\tfrac13,\tfrac13)\|_\infty \le 0.04.
\end{align}

\newcommand{\JudgeTabFont}{\small}

\definecolor{rowgray}{gray}{0.93}
\definecolor{sectiongray}{gray}{0.88}

\begin{table}[t]
\centering
\JudgeTabFont

\setlength{\tabcolsep}{4pt} 
\renewcommand{\arraystretch}{1.08}
\setlength{\extrarowheight}{0.6pt} 
\newcommand{\JudgeColGap}{\hspace{8pt}}

\begin{tabularx}{\columnwidth}{@{}>{\raggedright\arraybackslash}X r@{\JudgeColGap}r@{}}
\toprule
\rowcolor{sectiongray}
\textbf{Test} & \textbf{Independent} & \textbf{PBA} \\
\midrule

\rowcolor{sectiongray}
\multicolumn{3}{@{}l}{\textbf{Invariant} ($\mathcal{R}_{\mathrm{inv}}$)} \\
Aggrandising & 2 & \textbf{100} \\
Irrelevance & 2 & \textbf{98} \\
Repetition & 0 & \textbf{87} \\

\midrule
\rowcolor{sectiongray}
\multicolumn{3}{@{}l}{\textbf{Target Level Increase} ($\mathcal{R}_{\uparrow}(d,\ell)$)} \\
Low Evidence Addition & \textbf{92} & 90 \\
Medium Evidence Addition & 72 & \textbf{88} \\
High Evidence Addition & 95 & \textbf{97} \\

\midrule
\rowcolor{sectiongray}
\multicolumn{3}{@{}l}{\textbf{Target Level Decrease} ($\mathcal{R}_{\downarrow}(d,\ell)$)} \\
Low Evidence Debase & 73 & \textbf{85} \\
Medium Evidence Debase & \textbf{73} & 70 \\
High Evidence Debase & \textbf{95} & 92 \\

\midrule
\rowcolor{sectiongray}
\multicolumn{3}{@{}l}{\textbf{Other}} \\
Uniform Prior & \textbf{87} & 80 \\

\bottomrule
\end{tabularx}
\caption{Test pass rate by judge (rounded \%).}
\label{tab:test-pass-rate-by-judge}
\end{table}

\subsection{Defining Two Judges}
\label{sec:judge-selection}

We attempted to calibrate two LLM judges. Both were instructed to return the full set of posteriors in a single invocation.

\paragraph{Independent.}

At turn $t$, the judge analyses the full evidence set available at that turn and constructs posteriors directly:
\begin{align}
B_t^{\text{ind}}
=
\left\{
p(z_d \mid x_{0:t}, R)
\right\}_{d=1}^{D}.
\end{align}

\paragraph{Previous Belief Aware (PBA.)}

At turn $t$, the judge is also given the full evidence set $x_{0:t}$, but is additionally provided the previous belief $B_{t-1}^{\text{PBA}}$ and instructed to update by attending to how the evidence set changed from $x_{0:t-1}$ to $x_{0:t}$:
\begin{align}
B_t^{\text{PBA}}
=
\left\{
p\!\left(z_d \mid x_{0:t}, R, B_{t-1}^{\text{PBA}}, x_{0:t-1}\right)
\right\}_{d=1}^{D}.
\end{align}
The update is therefore \emph{change-aware} but still conditioned on the entire evidence set at turn $t$.

For initialisation (resume-only), we use a uniform prior:
\begin{align}
B_{-1}^{\text{PBA}}
=
\left\{
\left(\tfrac13,\tfrac13,\tfrac13\right)
\right\}_{d=1}^{D}
\end{align}

\subsection{Results}
\label{sec:judge-results}

\Cref{tab:test-pass-rate-by-judge} shows the calibration results. We can see that the PBA judge is highly stable relative to the Independent judge, as reasonably expected. This is crucial for belief convergence to be meaningful. Both judges generally perform similarly well across the other tests, but exhibit less inclination to remove probability mass from \medium despite performing well at \low and \high. This inertia seems reasonable given that \medium is the average, and more counter-evidence may be required to reduce evidential support of average compared to the extremes.

\section{Evaluating the System}
\label{sec:system-eval}

We now use simulation to demonstrate that our system can saturate a rubric with meaningful information about an applicant.

\subsection{Simulation}

\begin{table}[t!]
\centering
\small
\setlength{\tabcolsep}{4pt}
\begin{tabularx}{\columnwidth}{@{}>{\raggedright\arraybackslash}Xccc@{}}
\toprule
\textbf{Policy ($\pi$)} & \textbf{Resume} & \textbf{Rubric} & \shortstack{\textbf{Prev. belief}\\\textbf{$B_{t-1}$}} \\
\midrule
Belief Aware       & \cmark & \cmark & \cmark \\
Belief Unaware     & \cmark & \cmark & \xmark \\
Rubric Unaware     & \cmark & \xmark & \xmark \\
Shallow Unaware    & \cmark & \xmark & \xmark \\
\bottomrule
\end{tabularx}
\caption{Interviewer policies and their inputs at turn $t$. \textit{Shallow Unaware} is instructed to ask shallow, open-ended questions.}
\label{tab:interviewer-policies}
\end{table}

We firstly declare that this simulation serves as a coarse indicator of system behaviour rather than a rigorous test of interviewer strategy under the complexities and nuance of real interviews. The simulation components are:

\begin{compactitem}
    \item \textbf{Archetype:} Formally, an archetype \(a\) is a fixed latent rubric profile \(z^{(a)}_{1:D}\in\mathcal{L}^D\), where \(z^{(a)}_d\) is the archetype-implied latent level for dimension \(d\) (e.g.\ a $\mathrm{Tech\ Wizard}$ archetype might assign \(z^{(a)}_{\mathrm{programming}}=\high\) and \(z^{(a)}_{\mathrm{organisation}}=\low\)).

    \item \textbf{Judge:} We use our PBA judge as defined in \Cref{sec:judge-selection}.

    \item \textbf{Applicant:} We use an LLM to simulate applicant messages conditioned on a \textit{profile} (resume and archetype.) The LLM is provided the corresponding rubric to contextualise their archetype. Applicants are also seeded with a 1--2 word \textit{personality}.

    \item \textbf{Interviewer.} We also use an LLM for our interviewer and experiment with four policies, defined in \Cref{tab:interviewer-policies}. 
\end{compactitem}

Per rubric, we construct three bespoke archetypes, and three common \textit{anchor-archetypes} as defined by \Cref{eq:anchor-archetypes}, for a total of six:
\begin{equation}
\label{eq:anchor-archetypes}
\begin{aligned}
\text{low-logan}   &= \{z_d=\low\}_{d=1}^{D},\\
\text{average-jessie} &= \{z_d=\medium\}_{d=1}^{D},\\
\text{high-harley}  &= \{z_d=\high\}_{d=1}^{D}.
\end{aligned}
\end{equation}

Using the dataset defined in \Cref{sec:judge-test-dataset}, we take the product of 10 resumes and six archetypes per rubric domain for $P_\mathrm{full}=180$ total profiles across our three rubrics. For each profile, we initialise the judge with its resume and simulate $T=12$ interviewer-applicant turns. Due to practical realities, we expect that 12 responses is about the ceiling for what one could expect a real user to provide.

\subsection{Avoiding Applicant-Hacking}

A key risk was \textit{applicant-hacking}: the applicant leaking archetype cues \cite{lotfi-etal-2023-personalitychat} such that the judge can trivially infer their archetype. We sought to mitigate this via guardrails injected into the applicant prompt:
\begin{compactitem}
    \item \textbf{Self-preservation:} discourage explicit self-deprecation; require limitations to be implied rather than declared (esp.\ for \low KSA).
    \item \textbf{Show, don't tell:} elicit contextual evidence (examples/opinions) instead of listing traits verbatim.
    \item \textbf{No volunteered evidence:} answer only what is asked; do not add unrequested detail.
\end{compactitem}
We also used an LLM to sanitise the applicant's proposed response, without having seen the question, to reduce overreach of the applicant to its archetype's KSA, without being influenced by the interviewer.

\subsection{Choosing an Interviewer}

Given our calibrated PBA judge, we run a small-scale simulation on $P=30$ (randomly selected from the 180 profiles) and choose whichever causes the most belief change. We use turn-indexed cumulative posterior total variation (CTV) as a proxy for rubric-specific information gain accrued up to turn $t$. The per-turn posterior shift, $\Delta_t$, is defined in \Cref{eq:successive-tv}, and thus CTV at turn $t$ is defined as:
\begin{align}
\mathrm{CTV}_t(\pi)
=
\frac{1}{|\mathcal{P}|}
\sum_{p \in \mathcal{P}}
\sum_{u=1}^{t}
\Delta_{u}^{(p,\pi)},
\quad t \in \{1,\dots,T\}.
\label{eq:ctv}
\end{align}

\begin{figure}[t]
    \centering
    \includegraphics[width=\columnwidth]{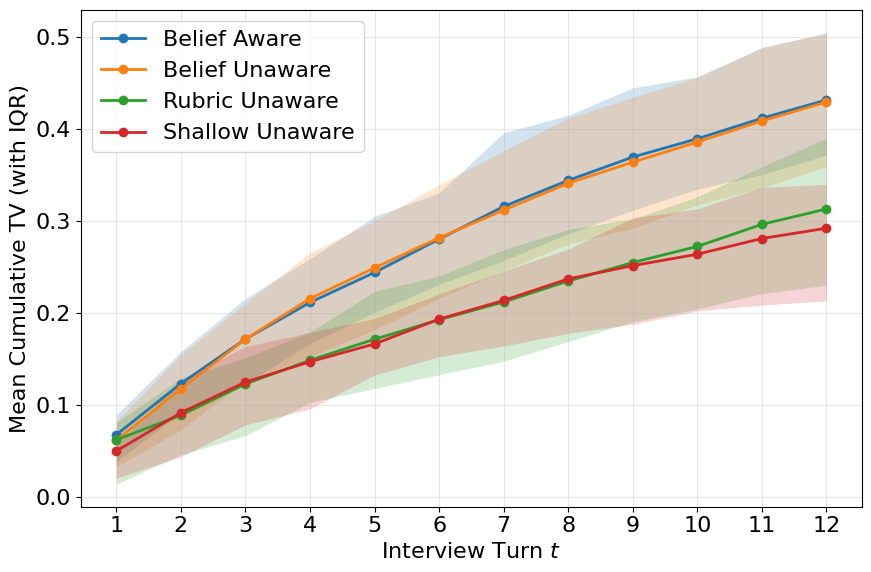}
    \caption{Comparison of \Cref{eq:ctv} over simulated interviews using a subset of $P=30$ profiles.}
    \label{fig:comparison-cumulative-tv}
\end{figure}

As can be seen in \Cref{fig:comparison-cumulative-tv}, all interviewer policies cause the judge to update its belief over the course of an interview and the inclusion of the rubric clearly discerns quality. Variation in interviewer quality is a strong argument against the existence of \textit{applicant-hacking}.
 
Another important observation is the little difference in performance between the \textit{Belief Aware} and \textit{Belief Unaware} policies. We suggest that the Belief Unaware interviewer is likely able to infer a reasonable belief state from the evidence alone given the relatively simplistic nature of the interviews. Acknowledging that we do not have an interviewer policy that meaningfully incorporates belief, we henceforth use the \textbf{Belief Unaware} policy for the full simulation and results in \Cref{sec:convergence} and \Cref{sec:archetype-recovery}.

\subsection{Demonstrating Convergence}
\label{sec:convergence}

\begin{figure}[t]
    \centering
    \includegraphics[width=\columnwidth]{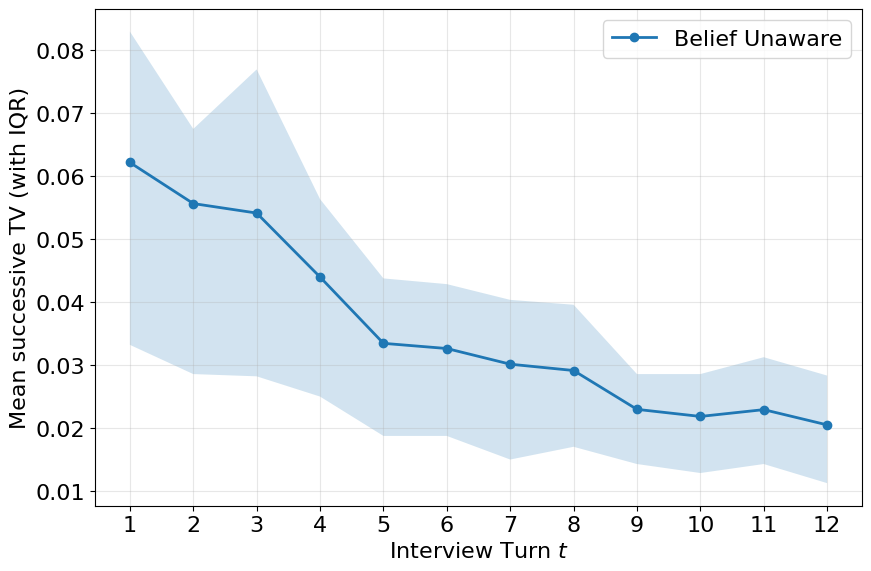}
    \caption{TV between judge's successive belief states $\Delta_t$ as described in \Cref{eq:successive-tv} over the course of $P_\mathrm{full}=180$ simulated interviews.}
    \label{fig:interviewer-convergence}
\end{figure}
We now revisit the concept of convergence introduced in \Cref{sec:interviewer-intro} as the reduction of TV between successive per-dimension posteriors in \Cref{eq:successive-tv}. \Cref{fig:interviewer-convergence} shows that the size of the judge's belief update (measured as TV) between $t=0$ (resume-only), and $t=1$ (resume + first turn), is $0.0621$ on average. The size of the judge's belief update at the last interview turn has reduced approximately $3\times$ to an average of $0.0205$, following a downward trend. This is strong evidence indicating that as our Belief Unaware interviewer elicits information from the applicant, the judge becomes increasingly confident in its belief over the applicant's KSA.

\subsection{Archetype Recovery}
\label{sec:archetype-recovery}

We now present that, as the judge's belief converges, it does so meaningfully, that is, towards an applicant's latent KSA. To do so, we report on \textit{archetype recovery} at the terminal turn \(T\). For each profile \(p\), we form a MAP rubric profile
\(\hat{z}_{1:D}^{(p)}\) from the final belief \(B_T^{(p)}\) via
\begin{equation}
\hat{z}_d^{(p)}=\arg\max_{\ell\in\mathcal{L}}[B_T^{(p)}]_d(\ell).
\end{equation}
Let \(M(\cdot)\) denote the standard one-hot encoding of a rubric profile into a
\(D\times L\) indicator matrix. We then predict the archetype by nearest-neighbour
matching over the archetype set \(\mathcal{A}\):
\begin{equation}
\hat{a}_T^{(p)}=\arg\min_{a\in\mathcal{A}}
\bigl\|M(\hat{z}_{1:D}^{(p)})-M(z_{1:D}^{(a)})\bigr\|_F.
\end{equation}
Recovery is the fraction of profiles for which \(\hat{a}_T^{(p)}=a^{(p)}\). Over \(P_{\mathrm{full}}=180\) we find that our system has an archetype recovery of $76.1\%$, compared to $16.7\%$ if recovery were to be performed at $t=0$ (resume-only). We observe that the most common level misclassifications are to an adjacent level, indicating that misclassifications at an extreme (i.e.\ \low or \high), still convey whether the latent level is \textit{below-average} or \textit{above-average}. See \Cref{sec:archetype-recovery-details} for further details.

\Cref{tab:model-comparison} (Appendix) provides a brief numerical comparison of the results in this section and \Cref{sec:convergence} with Google's Gemini 3.1 Pro \cite{gemini31pro_modelcard_2026}, which shows excellent performance as a potential alternative to GPT5.

\section{Related Work}

Automated hiring research has long emphasized resume-based screening and ranking, including competence-level estimation and machine-learning scoring of CVs \cite{Li2020CompetenceLevel,8117003,11147794}. In contrast, there is emerging work studying LLM-enabled interviewing systems, including a career pre-interview bot that dynamically generates slot-filling prompts \cite{hashimoto-etal-2025-career}, an NLP/ML competency interview bot grounded in behavioural event interviewing \cite{jokoarticle}, and a modular interviewer that generates expertise-aligned questions \cite{adeseye2025modular}. Commercial platforms similarly support early-stage evaluation and logistics. For example, HireVue and Modern Hire offer AI-enabled video interviewing and assessments.\footnote{\url{https://www.hirevue.com/}}\footnote{\url{https://www.modernhire.com/}} Since the inner-workings of such commercial models are opaque, our paper is the first to introduce an open-source system that incorporates rubric-aligned, auditable belief convergence.

\section{Conclusion and Future Work}

We propose that the system introduced by this paper can affordably proxy high-cost SME interviews at scale, thus improving the early-stage hiring process by better informing decision makers. We demonstrate, through simulation, that our calibrated judge's belief stabilises over time, and updates its belief in the direction of the simulated applicants' latent KSA, indicating that our system elicits meaningful information about the applicant.

In future work, we would like to study how to meaningfully incorporate belief in an interviewer policy using real data. For example, by using Bayesian Experimental Design to maximise information gain each turn \cite{choudhury2026bedllm}. Additionally, we would like to extend the system to ingest not only resumes, but also LinkedIn profiles, digital portfolios, and other sources provided by a candidate in a job application.

\clearpage

\section*{Limitations and Ethics}

\paragraph{Simulation}

We expect that our simulated applicants differ from human applicants. We would expect substantially more variation in a population of human applicants with respect to honesty, integrity, communication-ability, English-proficiency, conversation style, personality, etc. Further, our simulation is likely to miss the nuance and complexity inherent in interviewing a human applicant \cite{DBLP:conf/iui/ZhouWMYX19}, such as if the applicant were to ask clarifying questions, purport to be more proficient than they really are, or generally have a more complex and ambiguous latent KSA.

\paragraph{Reddit Dataset}

Although individuals who upload their data to \url{www.reddit.com} are making their data generally available to the public, we wished to anonymise all data used in our study. The r/Resumes subreddit rules request that users de-identify resumes before posting, as of 2026-02-27.\footnote{\url{https://www.reddit.com/r/Resumes/about/rules/}} We strived to obfuscate details and remove any remaining Personally Identifying Information (PII) regardless. We also provide only the extracted text with anonymised details in our released dataset. By not providing the raw PDF/image files, recovering which resume corresponds to which Reddit user is made to be a significantly more challenging task, further protecting any identities. We acknowledge that manual PII obfuscation may be imperfect.

\paragraph{Decision Empowering, not Making}

We strongly emphasise that we intend for our system to \textbf{empower} hiring personnel to make more informed decisions, not to make decisions on behalf of such personnel. It is broadly known that LLMs can make mistakes, and therefore, our system's judgements may be wrong. This may also be exacerbated by deliberate attempts to manipulate the judge's belief \cite{Maloyan2025Injoit, akdemir-levy-2025-resume-injections, mu2025aisecuritycoredomains, hwang-etal-2025-trick, SUEN2024123011}. We strongly encourage any instantiation of this work to be framed in the context of a ``human-in-the-loop'' system \cite{repec:eee:bushor:v:61:y:2018:i:4:p:577-586, Enarsson02012022, vats2025surveyhumanaicollaborationlarge}.

\paragraph{Safety and Bias}

Any production deployment of our system should involve safety guardrails to ensure that the interviewer does not ask about protected attributes (e.g., sex, ethnicity, etc.) We also acknowledge that LLMs can exhibit bias \cite{gallegos-etal-2024-bias, 10.1145/3597307}, especially towards proficiency in the English language \cite{10.1007/978-3-031-98417-4_6, hongli-etal-2024-mitigating} and thus further emphasize the importance of human verification and decision-making.

\bibliography{custom}

\clearpage

\appendix

\section{Judge Calibration Stratifications}
\label{sec:judge-strat}

We present selected stratifications in \Cref{tab:test-pass-rate-by-judge-stratified} for each calibration test class to highlight meaningful subsets. For the \textit{Invariant} tests, we stratify by input regimen (inclusion or exclusion of the stub, $\mathcal{I}$) and observe a slight performance reduction in \textit{Repetition} with $\mathcal{I}$. It is possible that $\mathcal{I}$ asserts a particular interview trajectory that is then disrupted by $\tau$, increasing the evidence deduplication burden. We stratify the \textit{Target Level Increase} tests by rubric domain and observe that the PBA judge struggles slightly with correctly recognising low evidence for the sales domain, and more-so with \medium evidence for the machine learning engineering domain. Lastly, we stratify the \textit{Target Level Decrease} tests by the type of debasement and observe that both judges struggle the most with \textit{Scope Downgrade}.

\paragraph{Defining Debasement.}

\begin{compactitem}
    \item \textbf{Context Invalidation:} the behaviour occurred, but is explained by legitimate constraints, role boundaries, appropriateness, or bandwidth, and is therefore not valid evidence of low capability or judgement.
    \item \textbf{Negative-Inference Clarification:} an earlier response \textit{sounded} low-capability, but subsequent clarification resolves ambiguity (wording/shorthand/oversimplification) and removes the negative inference.
    \item \textbf{Rigour Downgrade:} subsequent messages reveal the earlier evidence to be less systematic, deliberate, formal, or consistently applied than it initially implied.
    \item \textbf{Scope Downgrade:} subsequent messages narrow the applicant's personal ownership, responsibility, or contribution relative to what the earlier evidence suggested.
\end{compactitem}

\newcommand{\JudgeTabStratFont}{\small}

\definecolor{rowgray}{gray}{0.93}
\definecolor{sectiongray}{gray}{0.88}

\begin{table}[t]
\centering
\JudgeTabStratFont

\setlength{\tabcolsep}{4pt}     
\renewcommand{\arraystretch}{1.08} 
\setlength{\extrarowheight}{0.6pt} 
\newcommand{\JudgeColGap}{\hspace{8pt}}

\caption{Test pass rate by judge (rounded \%).}
\label{tab:test-pass-rate-by-judge-stratified}

\begin{tabularx}{\columnwidth}{@{}>{\raggedright\arraybackslash}X r@{\JudgeColGap}r@{}}
\toprule
\rowcolor{sectiongray}
\textbf{Test} & \textbf{Independent} & \textbf{PBA} \\
\midrule

\rowcolor{sectiongray}
\multicolumn{3}{@{}l}{\textbf{Invariant} ($\mathcal{R}_{\mathrm{inv}}$)} \\
\rowcolor{rowgray} Aggrandising & 2 & \textbf{100} \\
\hspace{0.9em}\textit{\(\hookrightarrow\) Without stub} & 3 & \textbf{100} \\
\hspace{0.9em}\textit{\(\hookrightarrow\) With stub} & 0 & \textbf{100} \\
\rowcolor{rowgray} Irrelevance & 2 & \textbf{98} \\
\hspace{0.9em}\textit{\(\hookrightarrow\) Without stub} & 0 & \textbf{100} \\
\hspace{0.9em}\textit{\(\hookrightarrow\) With stub} & 3 & \textbf{97} \\
\rowcolor{rowgray} Repetition & 0 & \textbf{87} \\
\hspace{0.9em}\textit{\(\hookrightarrow\) Without stub} & 0 & \textbf{90} \\
\hspace{0.9em}\textit{\(\hookrightarrow\) With stub} & 0 & \textbf{83} \\

\midrule
\rowcolor{sectiongray}
\multicolumn{3}{@{}l}{\textbf{Target Level Increase} ($\mathcal{R}_{\uparrow}(d,\ell)$)} \\
\rowcolor{rowgray} Low Evidence Addition & \textbf{92} & 90 \\
\hspace{0.9em}\textit{\(\hookrightarrow\) Machine Learning Engineering} & 85 & \textbf{95} \\
\hspace{0.9em}\textit{\(\hookrightarrow\) Graphic Design} & \textbf{100} & 95 \\
\hspace{0.9em}\textit{\(\hookrightarrow\) Sales} & \textbf{90} & 80 \\
\rowcolor{rowgray} Medium Evidence Addition & 72 & \textbf{88} \\
\hspace{0.9em}\textit{\(\hookrightarrow\) Machine Learning Engineering} & \textbf{75} & \textbf{75} \\
\hspace{0.9em}\textit{\(\hookrightarrow\) Graphic Design} & 65 & \textbf{100} \\
\hspace{0.9em}\textit{\(\hookrightarrow\) Sales} & 75 & \textbf{90} \\
\rowcolor{rowgray} High Evidence Addition & 95 & \textbf{97} \\
\hspace{0.9em}\textit{\(\hookrightarrow\) Machine Learning Engineering} & \textbf{100} & \textbf{100} \\
\hspace{0.9em}\textit{\(\hookrightarrow\) Graphic Design} & 90 & \textbf{100} \\
\hspace{0.9em}\textit{\(\hookrightarrow\) Sales} & \textbf{95} & 90 \\

\midrule
\rowcolor{sectiongray}
\multicolumn{3}{@{}l}{\textbf{Target Level Decrease} ($\mathcal{R}_{\downarrow}(d,\ell)$)} \\
\rowcolor{rowgray} Low Evidence Debase & 73 & \textbf{85} \\
\hspace{0.9em}\textit{\(\hookrightarrow\) Context invalidation} & 71 & \textbf{83} \\
\hspace{0.9em}\textit{\(\hookrightarrow\) Negative inference clarification} & 75 & \textbf{86} \\
\rowcolor{rowgray} Medium Evidence Debase & \textbf{73} & 70 \\
\hspace{0.9em}\textit{\(\hookrightarrow\) Rigour downgrade} & \textbf{75} & \textbf{75} \\
\hspace{0.9em}\textit{\(\hookrightarrow\) Scope downgrade} & \textbf{67} & 50 \\
\rowcolor{rowgray} High Evidence Debase & \textbf{95} & 92 \\
\hspace{0.9em}\textit{\(\hookrightarrow\) Rigour downgrade} & \textbf{97} & 91 \\
\hspace{0.9em}\textit{\(\hookrightarrow\) Scope downgrade} & \textbf{91} & \textbf{91} \\

\midrule
\rowcolor{sectiongray}
\multicolumn{3}{@{}l}{\textbf{Other}} \\
\rowcolor{rowgray} Uniform Prior & \textbf{87} & 80 \\

\bottomrule
\end{tabularx}
\end{table}

\section{Archetype Recovery Misclassifications}
\label{sec:archetype-recovery-details}

\begin{figure*}[t]
    \centering
    \includegraphics[width=\textwidth]{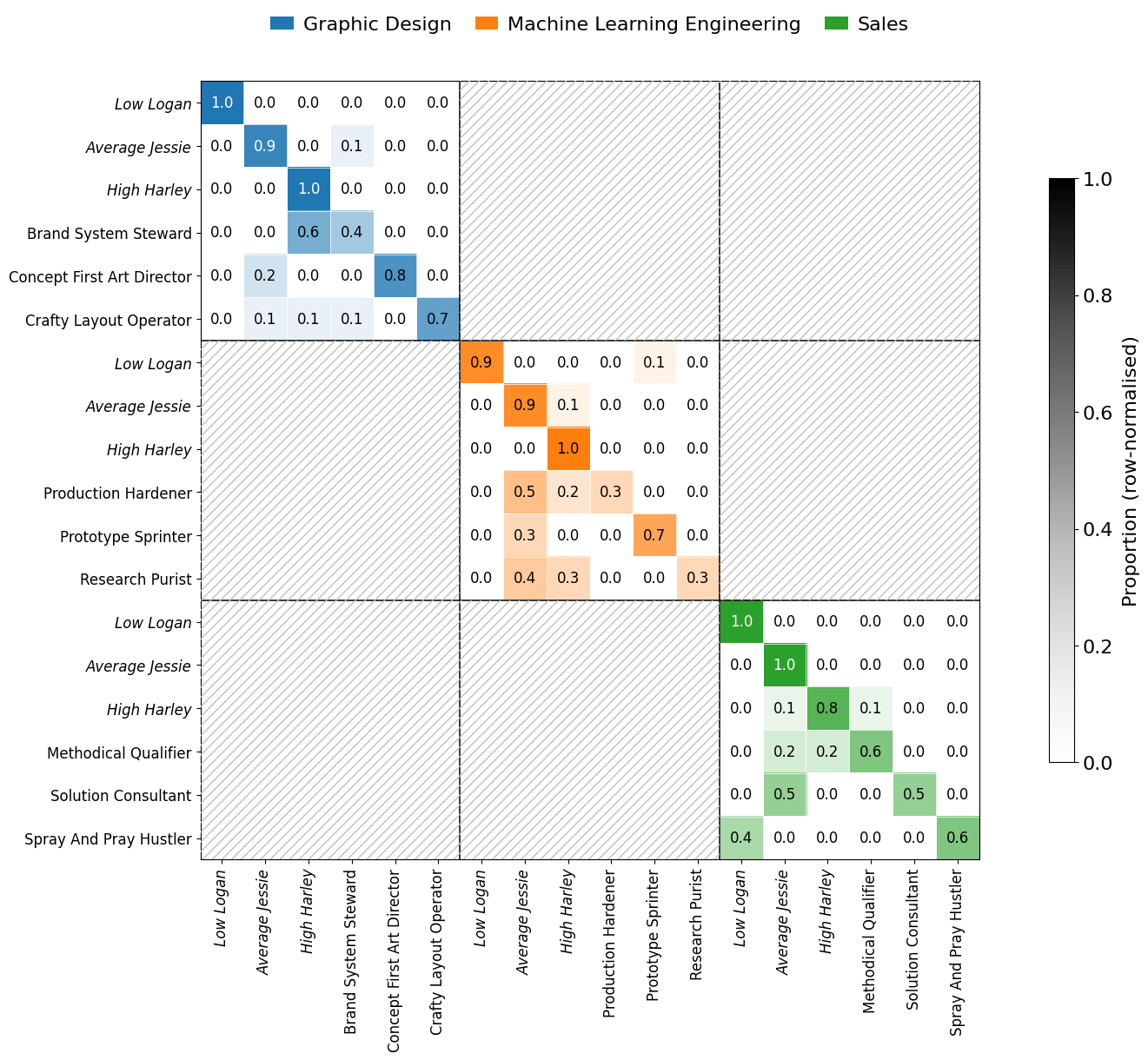}
    \caption{Confusion matrix for archetype recovery over the full simulation set ($|\mathcal{P}_{\mathrm{full}}|=180$). \textit{Italicised} archetypes are our previously defined anchor-archetypes \Cref{eq:anchor-archetypes}.}
    \label{fig:archetype-confusion}
\end{figure*}

To investigate misclassifications we plot a confusion matrix in \Cref{fig:archetype-confusion}. We observe that the judge struggles the most with recovering archetypes belonging to the \textit{Machine Learning Engineering} domain. This is consistent with our observation that the PBA judge is less proficient at aligning \medium evidence for Machine Learning Engineering compared to the other domains. It is worth noting that the Machine Learning Engineering rubric consists of only six dimensions, while \textit{Graphic Design} and \textit{Sales} consist of seven and eight respectively. While we do not assert a relationship between performance and the number of rubric dimensions, it stands to reason that if a rubric has a greater number of dimensions, it may tend to better separate KSA, and thus, better respect the \textit{internally-consistent} rubric requirement defined in \Cref{sec:rubric-def}. We also observe that misclassifications tend to select the anchor-archetypes defined in \Cref{eq:anchor-archetypes}. It is likely easier to gauge an applicant's \textit{overall} level than it is to spread belief across contrasting KSA.

\begin{figure}[t]
    \centering
    \includegraphics[width=\columnwidth]{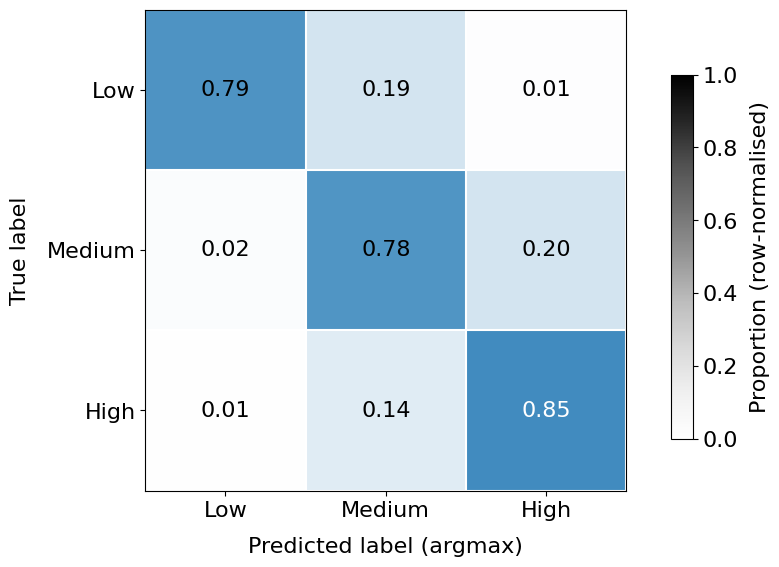}
    \caption{Per-level misclassifcations across the full simulation set ($P_{\mathrm{full}}=180$).}
    \label{fig:level-classifications}
\end{figure}

Finally, we examine the structure of these misclassifications at per-level granularity. Using the resume-only belief state ($t=0$), the normalised MAP-level frequencies across all profile-dimension pairs $(p,d)$ are $(f_{\low}, f_{\medium}, f_{\high}) = (0.39, 0.58, 0.04)$ (rounded). This indicates that resumes alone rarely support \high-level assignments, with most probability mass concentrated on \medium. Combined with the observation in \Cref{sec:judge-results} that the PBA judge is relatively inert in removing mass from \medium, this likely explains why misclassifications at either extreme predominantly collapse toward \medium, as shown in \Cref{fig:level-classifications}.

\begin{table}[t!]
\centering
\small
\setlength{\tabcolsep}{4pt}
\begin{tabularx}{\columnwidth}{@{}>{\raggedright\arraybackslash}Xcc@{}}
\toprule
\textbf{Model} & \textbf{TV ($\Delta_0 \rightarrow \Delta_T$)} & \textbf{AR ($0 \rightarrow T$)} \\
\midrule
GPT-5 & 0.0621 $\rightarrow$ 0.0205 & 16.7\% $\rightarrow$ 76.1\% \\
Gemini 3.1 Pro & 0.0640 $\rightarrow$ 0.0053 & 20.0\% $\rightarrow$ 100\% \\
\bottomrule
\end{tabularx}
\caption{Comparison of convergence and archetype recovery (AR) for GPT-5 vs.\ Gemini 3.1 Pro on $P=30$ profiles.}
\label{tab:model-comparison}
\end{table}

\end{document}